\documentclass[runningheads]{llncs}
\usepackage{graphicx}
\usepackage{subcaption}
\graphicspath{ {./images/} }
\usepackage{adjustbox}
\usepackage{hyperref}

\begin{document}
\title{Explainable Disease Classification via weakly-supervised segmentation}
\titlerunning{Explainable Disease Classification}
%
\author{Aniket Joshi, Gaurav Mishra \and Jayanthi Sivaswamy}
%
%
\institute{IIIT Hyderabad, India, 500032}
%
\maketitle              
\begin{abstract}
Deep learning based approaches to Computer Aided Diagnosis (CAD) typically pose the problem as an image classification (Normal or Abnormal) problem. These systems achieve high to very high accuracy in specific disease detection for which they are trained but lack in terms of an explanation for the provided decision/classification result. The activation maps which correspond to decisions do not correlate well with regions of interest for specific diseases.
This paper examines this problem and proposes an approach which mimics the clinical practice of looking for an evidence prior to diagnosis. A CAD model is learnt using a mixed set of information: class labels for the entire training set of images plus a rough localisation of suspect regions as an extra input for a smaller subset of training images for guiding the learning. The proposed approach is illustrated with detection of diabetic macular edema (DME) from OCT slices. Results of testing on on a large public dataset show that with just a third of images with roughly segmented fluid filled regions, the classification accuracy is on par with state of the art methods while providing a good explanation in the form of anatomically accurate heatmap /region of interest. The proposed solution is then adapted to Breast Cancer detection from mammographic images. Good evaluation results on public datasets underscores the generalisability of the proposed solution.

\end{abstract}
\section{Introduction}


Deep Learning (DL) based prediction systems which are black-boxes lack an explicit and declarative knowledge representation. Hence, despite their wide use for classification\cite{diabeticRetinoPathy,glaucomaDL,dekhil2017novel}, such systems have difficulty in generating the underlying explanatory structures\cite{Holzinger} which consequently impedes clinical adoption. Providing an evidence which might have led to the decision can mitigate this situation. In the case of diseases characterised by presence of lesions/abnormalities, assuming only image data is available, a natural option for this evidence is in the form of predicted regions of interest (ROI) which should be well aligned with locations of actual lesions/abnormalities(as annotated by experts).
\par
A common attempt towards explanation has been to use the activations of the last layer of the network used for the classification (using well known architectures such as Inception-V3, Resnet, AlexNet etc.) to get a heatmap which mark those regions in the image which might have led the model to give a particular prediction \cite{lee2019explainable,wang2017chestx}. Since the focus of such work is on classification, training is done with only labeled data and the reported evaluation is also of the classification accuracy and not of the heatmaps and their explanatory accuracy. Since these models were trained only on label information, there is no guarantee that the model will output a clinically accurate heatmap. An approach to get both class labels and accurate explanation would be to train the model using images which are annotated at both the image level and pixel/region/local level. 
This however poses a logistical challenge in the medical domain as labeled data are easier to extract from medical records and are available in abundance whereas region-level annotated data is not readily available to carry out a fully supervised segmentation.
\par
In this paper, we propose a novel approach to address the above problem. We propose a neural network architecture which can do both: learn to classify an image and leverage limited annotations to give an accurate heatmap. A novel training regime is designed to enable flexibility in the model building to accommodate and use varying levels of information that may be available.

\section{Method}
We illustrate our proposed method using Diabetic macular edema image classification in OCT image.
\subsection{Dataset(s)}OCT images (containing speckle noise) of retinal layers are used for DME detection. The fluid filled regions (FFR) in the retina can vary in size affecting the layer morphology.
A publicly available set \cite{zhangData} was chosen for our experiments. It has 84,495 OCT slices assigned one of 4 classes (NORMAL, CNV, DME and DRUSEN). We used the DME and NORMAL classes for our experiments. These images had no localizations for the fluid filled region. For generating the localizations, we trained a UNET model \cite{unet} on the other 2 datasets - 2015\_BOE\_Chiu\cite{chiu2015kernel} and RETOUCH\cite{RETOUCH} having 71 and 935 OCT slices with labelled segmentation maps for fluid filled regions respectively. This trained UNET model was used to predict the rough segmentation maps for dataset from \cite{zhangData} which was used later in all our experiments. A total of 16440 OCT slices (B-scans) were used in our experiments out of which 9332 were NORMAL cases and 7118 were DME affected. This was divided into Train, Validation and Test sets in the ratio 60:25:15. The normal images had empty segmentation masks while it represented the fluid filled regions in case of DME images.

Each of the OCT slice was preprocessed to identify the retinal layers which occupy only a small part of the OCT image. This was done using simple steps: row sum to vectorise the image and thresholding to extract the layered part.   

\subsection{Model}
The problem at hand is to build a model that is fully supervised for classification while also generating anatomically accurate heatmaps as an explanation for the class outputs using \textit{limited} region-level annotations. The assumption of limited availability of annotations constrains the approach to heatmap generation to be weakly supervised. Our solution is to design a network, that produces class label as a main output and an auxiliary output in the form of a rough segmentation map for a given input. Fig.\ref{my_model} shows the proposed network architecture for achieving this task. It consists of encoder and decoder segments, where the former is used to produce the classification output while the latter is used to generate the rough segmentation maps. 
A novel training methodology described in the next section ensures that the derived heatmaps using CAM \cite{cam} are not only guided by labels but also by the rough localisations.
Fewer filter layers (relative to a normal encoder) are employed at the end of the encoder for better localisation of the heatmap. After the last encoder layer (6x6x16 output), the output is flattened and a dense layer is added to obtain a class label as the output and heatmaps are derived using the method described in \ref{heatmaps}. 
 

\subsection{Training} \label{training}
Training of the model was done with images from the training set using $P$ images with only classification labels and $Q < P$ with both classification label and segmentation map. Thus, $Q:P$ indicates the proportion of different type of images used in training. 1:3 indicates that one third of the total number of training images had both classification label and segmentation map(See Table \ref{Data_specs}).  
\begin{table}[]
\centering
\caption{Public OCT data Specifications\cite{zhangData}}\label{tab1}
\begin{tabular}{|l|c|c|c|c|c|c|}
\hline
                                                  & \multicolumn{2}{c|}{Train}                             & \multicolumn{2}{c|}{Validation}                        & \multicolumn{2}{c|}{Test}                              \\ \hline
Data type                                         & \multicolumn{1}{l|}{Normal} & \multicolumn{1}{l|}{DME} & \multicolumn{1}{l|}{Normal} & \multicolumn{1}{l|}{DME} & \multicolumn{1}{l|}{Normal} & \multicolumn{1}{l|}{DME} \\ \hline
Images with Segmentation                          & 1866                        & 1425                     & 773                         & 583                      & 468                         & 364                      \\ \hline
\multicolumn{1}{|c|}{Images without Segmentation} & 3732                        & 2850                     & 1547                        & 1168                     & 936                         & 728                      \\ \hline
\end{tabular}
\label{Data_specs}
\end{table}

\begin{figure}[!htb]
    \centering
    \includegraphics[width=\textwidth]{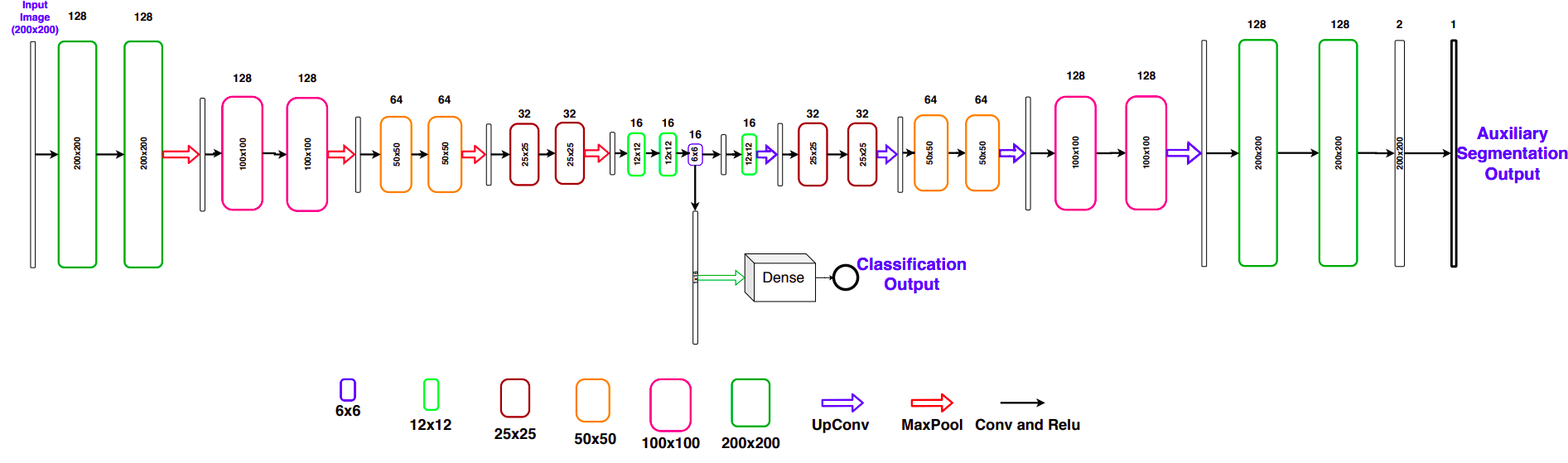}
    \caption{Proposed network architecture with 2 outputs - Classification and Auxiliary Segmentation. Left Branch before Classification output is called ED branch.}
    \label{my_model}
\end{figure}

A special training regime is designed for the given task. The model has two branches (See Fig.\ref{my_model} with two outputs, namely, the class label and a rough segmentation map). The left branch (encoder before the classification output) together with the dense layers is a unit (referred to as ED) whose output is the class label. 

\par 
In the first phase, the ED unit is trained for a few (5) epochs following which the whole model is trained for some (10) epochs. At the end of this phase, the ED part of the network will learn the features aiding both classification and segmentation tasks. In the second phase, only the dense layers of ED unit are trained keeping all the other weights constant (for 20 epochs). This second phase forces the model to predict the class label using the features that were learned during the whole model training for segmentation and classification. This also serves to boost the classification accuracy of the model. In the third and final phase of training, the whole model is trained again so that the encoder part learns more of the features that will be used for segmentation (for 50 epochs). This phase of training is made to be the longest so that the model is forced to learn more of the segmentation features so as to get the best heatmaps using the regime described in Section\ref{heatmaps}.
\par 
In terms of loss functions, binary cross entropy loss is used while training only the ED unit of the model whereas while training the entire model, a weighted sum of dice coefficient loss and binary cross entropy loss is used. The weight is the hyper-parameter and is taken as 1 during the training phase. Higher weight for dice coefficient loss will give us better region of interests(heatmaps) but at the cost of lower classification accuracy.

\subsection{Deriving the heatmaps} \label{heatmaps}
\begin{figure}
\centering
  \begin{subfigure}[b]{0.192\textwidth}
    \includegraphics[width=\textwidth, height=2.0cm]{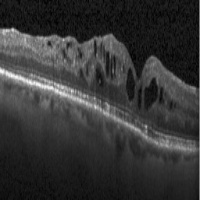}
    \label{fig:1}
  \end{subfigure}
  \begin{subfigure}[b]{0.192\textwidth}
    \includegraphics[width=\textwidth, height=2.0cm]{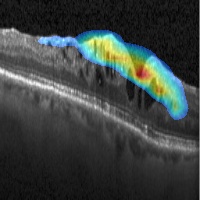}
    \label{fig:2}
  \end{subfigure}
    \begin{subfigure}[b]{0.192\textwidth}
    \includegraphics[width=\textwidth, height=2.0cm]{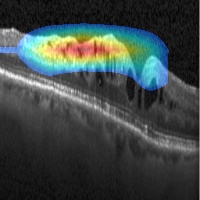}
    \label{fig:2}
  \end{subfigure}
    \begin{subfigure}[b]{0.192\textwidth}
    \includegraphics[width=\textwidth, height=2.0cm]{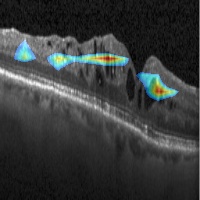}
    \label{fig:2}
  \end{subfigure}
    \begin{subfigure}[b]{0.192\textwidth}
    \includegraphics[width=\textwidth, height=2.0cm]{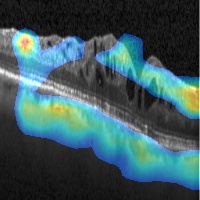}
    \label{fig:1}
  \end{subfigure}
  \begin{subfigure}[b]{0.192\textwidth}
    \includegraphics[width=\textwidth, height=2.0cm]{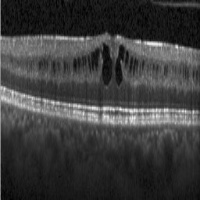}
    \label{fig:1}
  \end{subfigure}
  \begin{subfigure}[b]{0.192\textwidth}
    \includegraphics[width=\textwidth, height=2.0cm]{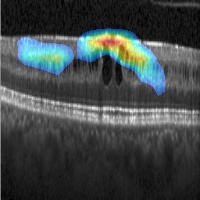}
    \label{fig:2}
  \end{subfigure}
    \begin{subfigure}[b]{0.192\textwidth}
    \includegraphics[width=\textwidth, height=2.0cm]{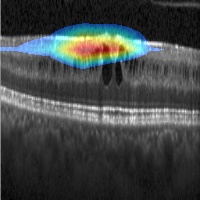}
    \label{fig:2}
  \end{subfigure}
    \begin{subfigure}[b]{0.192\textwidth}
    \includegraphics[width=\textwidth, height=2.0cm]{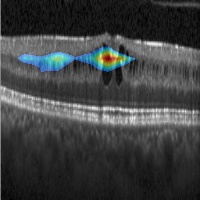}
    \label{fig:2}
  \end{subfigure}
  \begin{subfigure}[b]{0.192\textwidth}
    \includegraphics[width=\textwidth, height=2.0cm]{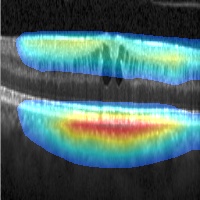}
    \label{fig:1}
  \end{subfigure}
  \begin{subfigure}[b]{0.192\textwidth}
    \includegraphics[width=\textwidth, height=2.0cm]{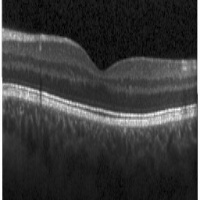}
    \caption{Original}
    \label{fig:1}
  \end{subfigure}
  \begin{subfigure}[b]{0.192\textwidth}
    \includegraphics[width=\textwidth, height=2.0cm]{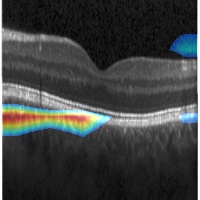}
    \caption{ROI 1:1}
    \label{fig:2}
  \end{subfigure}
    \begin{subfigure}[b]{0.192\textwidth}
    \includegraphics[width=\textwidth, height=2.0cm]{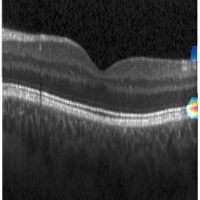}
    \caption{ROI 1:2}
    \label{fig:2}
  \end{subfigure}
    \begin{subfigure}[b]{0.192\textwidth}
    \includegraphics[width=\textwidth, height=2.0cm]{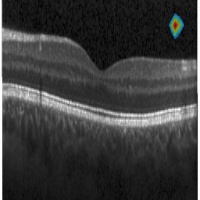}
    \caption{ROI 1:3}
    \label{fig:2}
  \end{subfigure}
      \begin{subfigure}[b]{0.192\textwidth}
    \includegraphics[width=\textwidth, height=2.0cm]{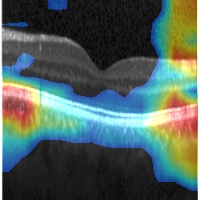}
    \caption{ROI $M_b$}
    \label{fig:1}
  \end{subfigure}
  \caption{DME detection. Sample ROI outputs for training with different ratios of labeled images and local annotation for DME (rows 1,2) and normal (row 3) cases.}
  \label{ROI_ratio}
\end{figure}

The desired heatmaps (as explanation for the predicted class label) were derived using the following steps. The dense layers from the ED unit were removed and a global average pooling layer was added to the final encoder layer (6x6x16 output) to give a 1x16 feature map. The dense layer was attached finally to give the classification output. 
Only the 16 neurons of the dense layer were trained in this phase and all the other weights were kept constant from the last phase of the training. Heatmaps were obtained using the Class Activation Mapping (CAM)\cite{cam}. 
As the convolution layer weights were taken from the last phase of the training, it had the features which were learned for both segmentation and classification. In our case, the adapted CAM approach was weakly supervised, as opposed to the normal CAM where heatmaps are simply an extracted by-product of the main classification task. Using the adapted CAM, model design and training methodology, it was ensured that the heatmaps are more accurately localised then the normal CAM approach whose results can be seen in the last column($M_b$) and last row of Fig.\ref{ROI_ratio} and Table \ref{results_ratio} respectively.

\section{Experiments}
Several experiments were done to assess the proposed idea for generating explainable classification. Here, ED unit upon training with only the classification labelled data forms the base classification model, i.e. $M_b$ with $Q=0$. In the first experiment, training was done with different values for 1:R, R=1,2,3,4 to assess the effect of lowering the value of $Q$ on the performance, namely accuracy of classification and the generated heatmaps. 
In the second experiment, we wished to understand if the degree of accuracy of local annotation affects the model's performance.  Training was done with 4 types of annotations: roughly accurate segmentation boundary of each FFR; a bounding box for each FFR; randomly generated image patches and finally the whole image (i.e. no localisation at all). Quantitative assessment of the classification task is done using Accuracy, Sensitivity, Specificity and AUC. 
Correctness of ROI prediction for a particular image is accessed using a method described later and the accuracy of detection which is the number of images in which ROIs are predicted correctly to the total number of images is reported in \%. 

\begin{table}[]
\caption{DME detection results with different ratios of local annotation}
\begin{adjustbox}{width=\textwidth}
\begin{tabular}{|c|c|c|c|c|c|c|c|}

\hline
   Training                  & \multicolumn{4}{c|}{Classification}           & \multicolumn{3}{c|}{ROI Prediction}    \\ \hline
 data ratio 1:R & Accuracy & AUC    & Specificity & Sensitivity & Total Images & Images with correct ROIs & Accuracy of detection \\ \hline
1:1                 & 91.38    & 96.56 & 94.65       & 87.17       & 993          & 951          & \textbf{95.7}     \\ \hline
1:2                 & 92.06    & 96.39 & 95.44       & 87.72       & 1004         & 947          & 94.3     \\ \hline
1:3                 & 97.71    & 98.95 & 98.57       & 96.61       & 1056         & \textbf{978}          & 92.6     \\ \hline
1:4                 & \textbf{97.86}    & 98.95 & \textbf{98.6}        & 96.93       & 1061         & 610            & 57.49        \\ \hline
$M_b$             & 97.83     & \textbf{99.27}  & 97.50       & \textbf{98.26}       & 1061
   & 395          & 37.2    \\ \hline
\end{tabular}
\end{adjustbox}
\label{results_ratio}
\end{table}

\begin{figure}[!htb]
\centering
  \begin{subfigure}[b]{0.24\textwidth}
    \includegraphics[width=\textwidth, height=2.0cm]{images/1.jpeg}
    \label{fig:1}
  \end{subfigure}
  \begin{subfigure}[b]{0.24\textwidth}
    \includegraphics[width=\textwidth, height=2.0cm]{images/1_3_1.jpeg}
    \label{fig:2}
  \end{subfigure}
    \begin{subfigure}[b]{0.24\textwidth}
    \includegraphics[width=\textwidth, height=2.0cm]{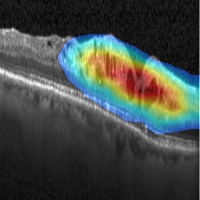}
    \label{fig:2}
  \end{subfigure}
    \begin{subfigure}[b]{0.24\textwidth}
    \includegraphics[width=\textwidth, height=2.0cm]{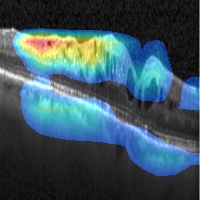}
    \label{fig:2}
  \end{subfigure}
  \begin{subfigure}[b]{0.24\textwidth}
    \includegraphics[width=\textwidth, height=2.0cm]{images/2.jpeg}
    \caption{Original}
    \label{fig:1}
  \end{subfigure}
  \begin{subfigure}[b]{0.24\textwidth}
    \includegraphics[width=\textwidth, height=2.0cm]{images/1_3_2.jpeg}
    \caption{Accurate map}
    \label{fig:2}
  \end{subfigure}
    \begin{subfigure}[b]{0.24\textwidth}
    \includegraphics[width=\textwidth, height=2.0cm]{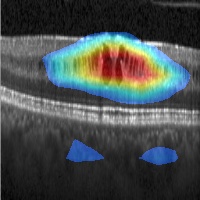}
    \caption{Bounding box}
    \label{fig:2}
  \end{subfigure}
    \begin{subfigure}[b]{0.24\textwidth}
    \includegraphics[width=\textwidth, height=2.0cm]{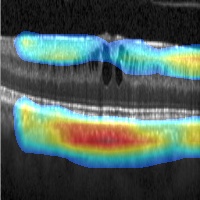}
    \caption{Whole image}
    \label{fig:2}
  \end{subfigure}
  \caption{ROI obtained for training with different types of local annotation.}
  \label{roi_localizations}
\end{figure}

\section{Results}
The ROIs derived with models trained using different ratios (1:R) of images with labels and with labels + localisation annotation are shown for 3 sample test images in Fig.\ref{ROI_ratio}. 
The results for $M_b$ (last column) are diffuse ROI covering almost the entire image. With the addition of more and more images with localisation information during the training phase, the ROIs improve progressively and we get the best overlap with the FFRs when R reaches 1.
The intersection over union (IOU) metric was used to help quantitatively assess the derived ROIs against the ground truth for FFR. An IOU threshold of 0.3 is taken to declare correct detection of ROI. Table \ref{results_ratio} lists the number of correctly detected ROIs and the accuracy of detection(Correct ROIs/Total Images). These results are consistent with the qualitative results showing an increasing trend in accuracy of detection as R value approaches 1. Lowest ROI detection accuracy is obtained by $M_b$ due to lack of information about suspect regions during training. 

\begin{figure}[!htb]
\centering
  \begin{subfigure}[b]{0.19\textwidth}
    \includegraphics[width=\textwidth, height=2.0cm]{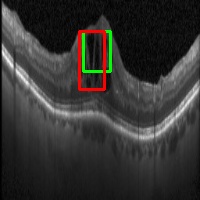}
    \label{fig:1}
  \end{subfigure}
  \begin{subfigure}[b]{0.19\textwidth}
    \includegraphics[width=\textwidth, height=2.0cm]{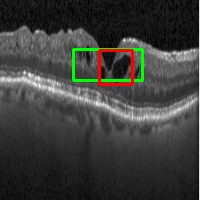}
    \label{fig:2}
  \end{subfigure}
    \begin{subfigure}[b]{0.19\textwidth}
    \includegraphics[width=\textwidth, height=2.0cm]{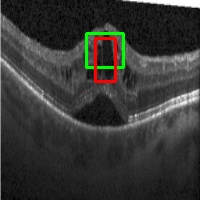}
    \label{fig:2}
  \end{subfigure}
    \begin{subfigure}[b]{0.19\textwidth}
    \includegraphics[width=\textwidth, height=2.0cm]{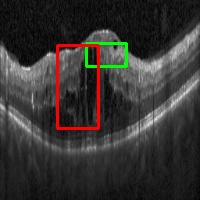}
    \label{fig:2}
  \end{subfigure}
     \begin{subfigure}[b]{0.19\textwidth}
    \includegraphics[width=\textwidth, height=2.0cm]{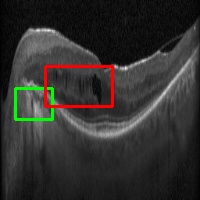}
    \label{fig:2}
  \end{subfigure}
   \caption{Predicted ROI (in green) for DME detection. Ground Truth is in red. }
 \label{roi_bounding_box}
\end{figure}

Next, we present the results of experimenting with different types of localization of suspect regions during training. Fig.\ref{roi_localizations} shows the derived ROIs for 2 sample images. It can be seen that ROI is less and less localised as the precision with which local annotations used in training data is compromised, which is to be expected.  Quantitatively, accuracy of classification remains above 90 \% for all types of local annotations used (see Table \ref{seg}). However, there is a fall in accuracy of correct ROI predictions for a bounding box type of annotation and a steep degradation when the whole image or random patches are used as annotations. More results are shown in Fig.\ref{roi_bounding_box} for 1:3 training regime.
A comparison of the classification accuracy of the proposed method with 2 state of the art (SOTA) methods are given in Table \ref{sota}. Our method is seen to be almost on par with \cite{cellPaper} for 4 metrics, when tested on the same large dataset \cite{zhangData}.

\begin{table}[]
\centering
\caption{DME detection. Results with different types of local annotations.}
\begin{adjustbox}{width=\textwidth}
\begin{tabular}{|c|c|c|c|c|c|c|c|}
\hline
                  Annotation   & \multicolumn{4}{c|}{Classification}          & \multicolumn{3}{c|}{ROI Prediction}    \\ \hline
Type & Accuracy & AUC   & Specificity & Sensitivity & Total Images & Correct ROIs & Accuracy of detection \\ \hline
Accurate Map        & \textbf{97.71}    & 98.95 & \textbf{98.57}       & \textbf{96.61}       & 1056         & \textbf{978}          & \textbf{92.6}    \\ \hline
Bounding Box        & 94.91    & \textbf{99.01} & 96.86       & 92.39       & 1035         & 803          & 77.5    \\ \hline
Whole Image         & 93.38    & 98.32 & 95.01       & 91.30       & 1019         & 345          & 33.8    \\ \hline
Random              & 92.42    & 97.49 & 94.65       & 90.56       & 1010         & 432          & 42.77   \\ \hline
\end{tabular}
\end{adjustbox}
\label{seg}
\end{table}

\begin{table}[]
\centering
\caption{DME detection performance comparison with SOTA}
\begin{tabular}{|c|c|c|c|c|c|}
\hline
Method                   & Dataset                                             & Accuracy & AUC   & Specificity & Sensitivity \\ \hline
Kermany et. al{\cite{cellPaper}}   & \cite{zhangData}                                               & 98.2     & 99.87 & 99.6        & 96.8        \\ \hline
\textbf{Our method(1:3)}          & \cite{zhangData}                                               & 97.71    & 98.95 & 98.57       & 96.61       \\ \hline
Srinvasan et. al{\cite{srinivasan}} & Duke                                                & 93.335   & -     & 93.8        & 68.8        \\ \hline
\end{tabular}
\label{sota}
\end{table}

\begin{figure}[!htb]
\centering
  \begin{subfigure}[b]{0.225\textwidth}
    \includegraphics[width=\textwidth, height = 3.5cm]{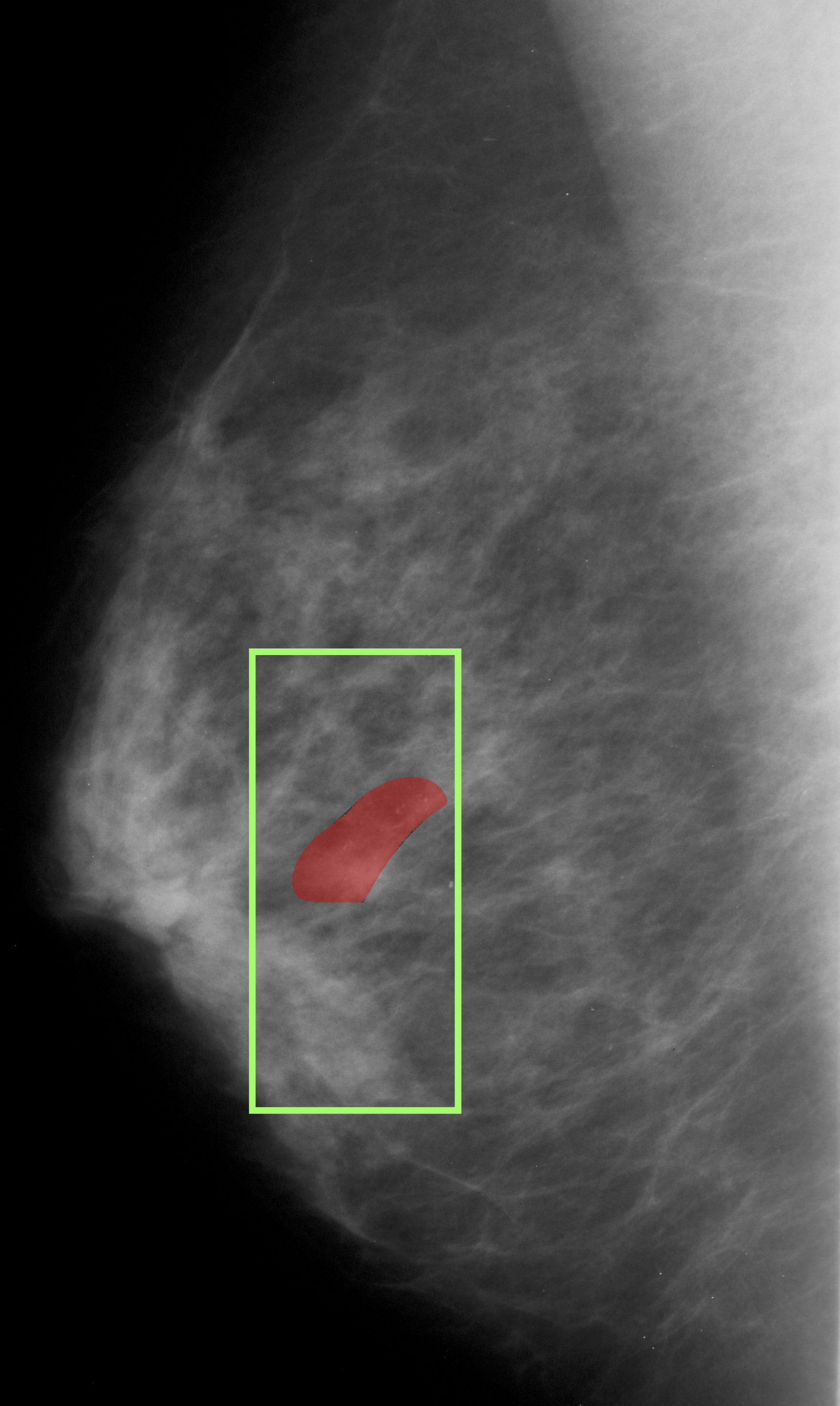}
    \label{fig:1}
  \end{subfigure}
  \begin{subfigure}[b]{0.225\textwidth}
    \includegraphics[width=\textwidth, height = 3.5cm]{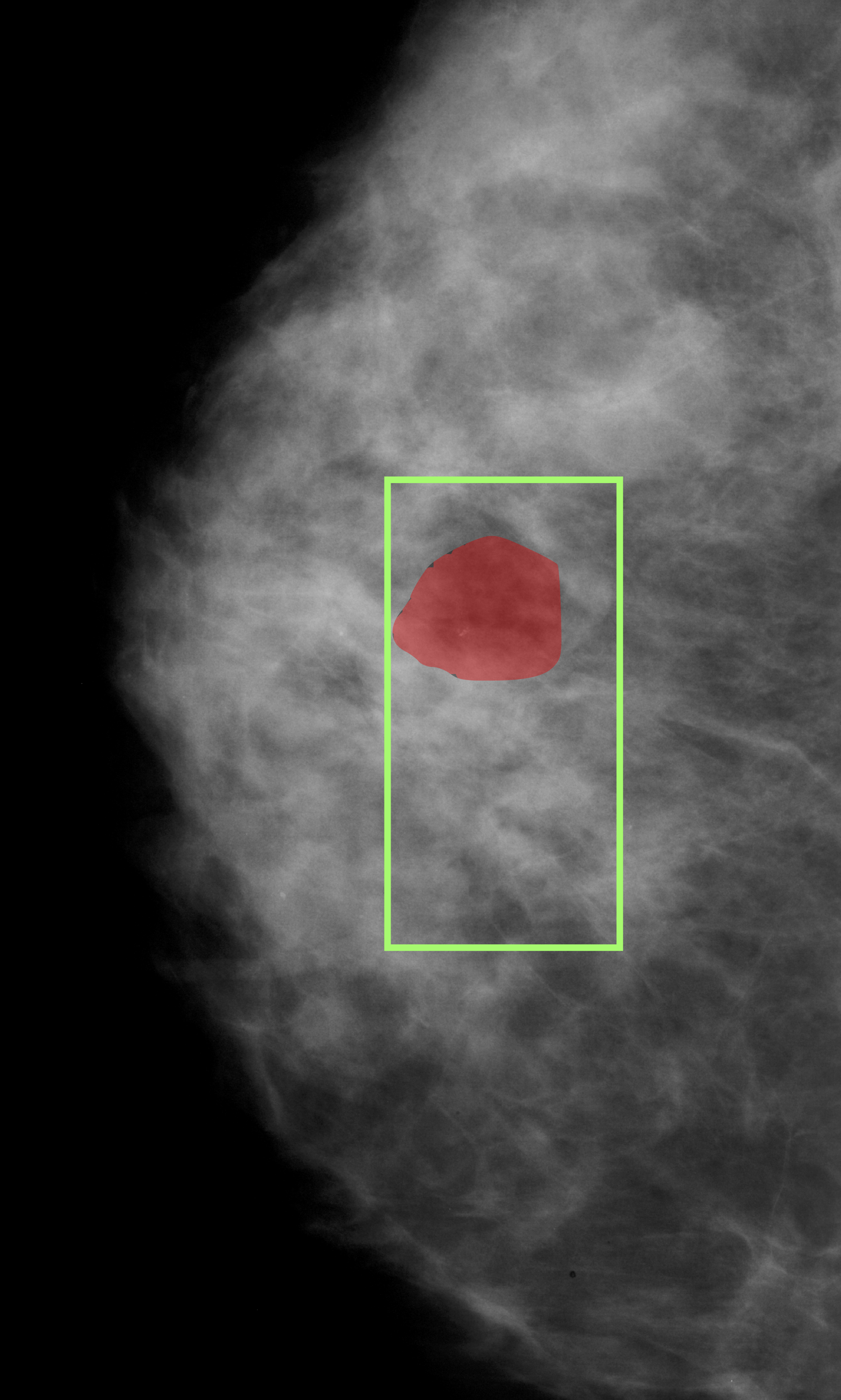}
    \label{fig:2}
  \end{subfigure}
  \begin{subfigure}[b]{0.225\textwidth}
    \includegraphics[width=\textwidth, height = 3.5cm]{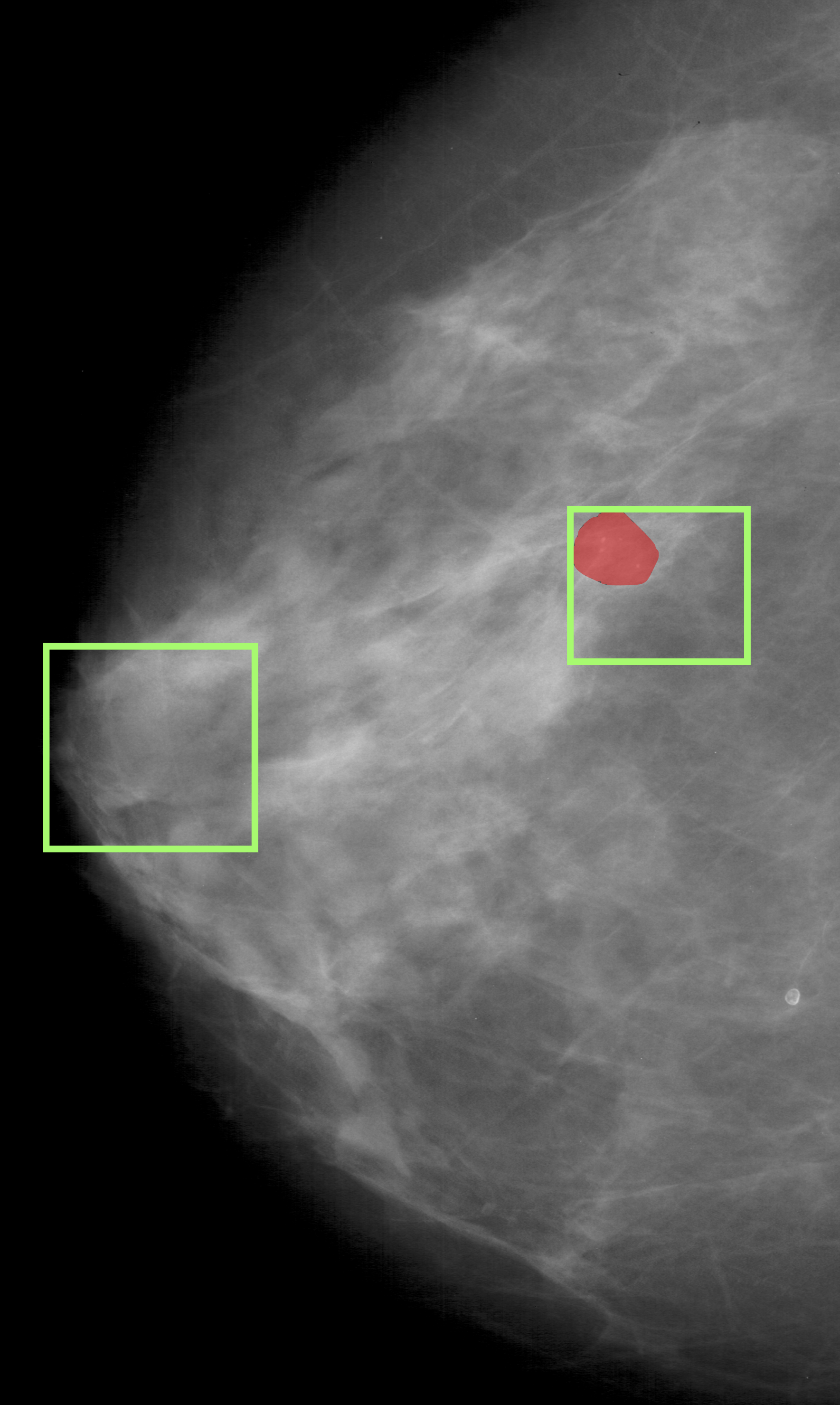}
    \label{fig:1}
  \end{subfigure}
 \caption{Breast cancer detection. Green - Predicted ROI, Red - Ground Truth. }
 \label{mamogram_outputs}
\end{figure}

The proposed approach was also applied to breast cancer detection (for screening) from mammograms. Unlike the DME problem, evidence for breast cancer is not easily discernible to the naked, untrained eye and hence is particularly challenging. 
The 2 classes of interest were defined to be  normal and abnormal. The latter includes benign and malignant cases as discrimination between these cases is difficult and best done by a specialised model. The ROI prediction aimed at are suspect regions regardless of whether they are benign or malignant. The architecture used for the DME problem was used and patches from the entire mammogram was fed as input. Training methodology was as described in section \ref{training}. The patches(200x200) which were classified as positive by the model acts as the predicted ROI in the large sized mammogram image(around 4000x4000 in dimension). The model was assessed on CBIS-DDSM dataset \cite{cbisddsm}. A total of 5218 training images (2017 abnormal and 3201 Normal) were split into train and validation in the ratio of 12:5. A model trained on 1:3 ratio of annotated images, was evaluated on a test set of 1298 images(709 abnormal and 589 Normal). The AUC/sensitivity(SN)\%/specificity(SP)\% attained was 0.98/90/93 respectively. Three sample images with ground truth regions and model-predicted ROIs(bounding boxes) are shown in Fig.\ref{mamogram_outputs}. A baseline model($M_b$) was also trained and tested. It achieved a AUC/SN/SP of 0.972/88.2/ 91.3 respectively. A recent method \cite{shen2019deep} that does normal/cancerous classification also reports on \cite{cbisddsm}. It is based on transfer learning with a Resnet50 and reports AUC/SN/SP to be 0.91/86\%/80.1\%.

\section{Conclusion and Discussion}
The need to make decisions of CAD systems for disease detection more explainable was addressed in this paper. It is worth emphasising that the primary problem here is not segmentation and that the evidence provided are an added benefit to the main classification task. Using a data set with very limited local annotations, a lightweight network design was proposed and trained using a novel methodology to provide classification and an explanation via heatmaps. The approach has been illustrated for DME and breast cancer detection, both employing different modality. 
The proposed solution also serves to draw the attention of the image reader to the areas deemed to be suspect. Results of extensive experiments indicate that a model trained with labeled images where only a third have basic bounding box type of local annotation, can achieve above 90\% classification accuracy and provide explanation in the form of heatmaps with over 75\% accuracy; the accuracy of heatmaps do improve with more accurate and abundant local annotation. The proposed solution thus enables an explainable CAD design with a \textit{flexible} use of available annotations.
\bibliographystyle{ieeetr}
\bibliography{refer}

\begin{thebibliography}{10}

\bibitem{diabeticRetinoPathy}
J.~Sahlsten, J.~Jaskari, J.~Kivinen, L.~Turunen, E.~Jaanio, K.~Hietala, and
  K.~Kaski, ``Deep learning fundus image analysis for diabetic retinopathy and
  macular edema grading,'' {\em Scientific reports}, vol.~9, no.~1, pp.~1--11,
  2019.

\bibitem{glaucomaDL}
X.~Chen, Y.~Xu, D.~W.~K. Wong, T.~Y. Wong, and J.~Liu, ``Glaucoma detection
  based on deep convolutional neural network,'' in {\em 2015 37th annual
  international conference of the IEEE engineering in medicine and biology
  society (EMBC)}, pp.~715--718, IEEE, 2015.

\bibitem{dekhil2017novel}
O.~Dekhil, M.~Ismail, A.~Shalaby, A.~Switala, A.~Elmaghraby, R.~Keynton,
  G.~Gimel'farb, G.~Barnes, and A.~El-Baz, ``A novel cad system for autism
  diagnosis using structural and functional {MRI},'' in {\em 2017 IEEE ISBI},
  pp.~995--998, IEEE, 2017.

\bibitem{Holzinger}
A.~Holzinger, C.~Biemann, C.~S. Pattichis, and D.~B. Kell, ``What do we need to
  build explainable {AI} systems for the medical domain?,'' {\em arXiv preprint
  arXiv:1712.09923}, 2017.

\bibitem{lee2019explainable}
H.~Lee, S.~Yune, M.~Mansouri, M.~Kim, S.~H. Tajmir, C.~E. Guerrier, S.~A.
  Ebert, S.~R. Pomerantz, J.~M. Romero, S.~Kamalian, {\em et~al.}, ``An
  explainable deep-learning algorithm for the detection of acute intracranial
  haemorrhage from small datasets,'' {\em Nature Biomedical Engineering},
  vol.~3, no.~3, p.~173, 2019.

\bibitem{wang2017chestx}
X.~Wang, Y.~Peng, L.~Lu, Z.~Lu, M.~Bagheri, and R.~M. Summers, ``Chestx-ray8:
  Hospital-scale chest x-ray database and benchmarks on weakly-supervised
  classification and localization of common thorax diseases,'' in {\em
  Proceedings of the IEEE conference on computer vision and pattern
  recognition}, pp.~2097--2106, 2017.

\bibitem{zhangData}
D.~Kermany, K.~Zhang, and M.~Goldbaum, ``Large dataset of labeled optical
  coherence tomography (oct) and chest x-ray images,'' 2018.

\bibitem{unet}
O.~Ronneberger, P.~Fischer, and T.~Brox, ``U-net: Convolutional networks for
  biomedical image segmentation,'' in {\em International Conference on MICCAI},
  pp.~234--241, Springer, 2015.

\bibitem{chiu2015kernel}
S.~J. Chiu, M.~J. Allingham, P.~S. Mettu, S.~W. Cousins, J.~A. Izatt, and
  S.~Farsiu, ``Kernel regression based segmentation of optical coherence
  tomography images with diabetic macular edema,'' {\em Biomedical optics
  express}, vol.~6, no.~4, pp.~1172--1194, 2015.

\bibitem{RETOUCH}
H.~Bogunovi\'c, F.~Venhuizen, S.~Klimscha, S.~Apostolopoulos, A.~Bab-Hadiashar,
  U.~Bagci, {\em et~al.}, ``{RETOUCH - The Retinal OCT Fluid Detection and
  Segmentation Benchmark and Challenge},'' {\em IEEE Transactions on Medical
  Imaging}, vol.~38, pp.~1858--1874, Aug 2019.

\bibitem{cam}
B.~Zhou, A.~Khosla, A.~Lapedriza, A.~Oliva, and A.~Torralba, ``Learning deep
  features for discriminative localization,'' in {\em Proceedings of the IEEE
  conference on CVPR}, pp.~2921--2929, 2016.

\bibitem{cellPaper}
D.~S. Kermany, M.~Goldbaum, W.~Cai, C.~C. Valentim, H.~Liang, S.~L. Baxter,
  A.~McKeown, G.~Yang, X.~Wu, F.~Yan, {\em et~al.}, ``Identifying medical
  diagnoses and treatable diseases by image-based deep learning,'' {\em Cell},
  vol.~172, no.~5, pp.~1122--1131, 2018.

\bibitem{srinivasan}
P.~Srinivasan, L.~Kim, P.~Mettu, S.~Cousins, G.~Comer, J.~Izatt, and S.~Farsiu,
  ``Fully automated detection of diabetic macular edema and dry age-related
  macular degeneration from optical coherence tomography images,'' {\em
  Biomedical Optics Express}, vol.~5, 10 2014.

\bibitem{cbisddsm}
R.~S. Lee, F.~Gimenez, A.~Hoogi, K.~K. Miyake, M.~Gorovoy, and D.~L. Rubin, ``A
  curated mammography data set for use in computer-aided detection and
  diagnosis research,'' {\em Scientific data}, vol.~4, p.~170177, 2017.

\bibitem{shen2019deep}
L.~Shen, L.~R. Margolies, J.~H. Rothstein, E.~Fluder, R.~McBride, and W.~Sieh,
  ``Deep learning to improve breast cancer detection on screening
  mammography,'' {\em Scientific reports}, vol.~9, no.~1, pp.~1--12, 2019.

\end{thebibliography}


\end{document}